\documentclass[letterpaper, 10 pt, conference]{Packages/ieeeconf}  
\usepackage{microtype}  
\microtypesetup{protrusion=true,expansion=true}
\IEEEoverridecommandlockouts                              
 
\usepackage{float}
\usepackage{times}
\usepackage{epsfig}
\usepackage{graphicx}
\usepackage{amsmath}
\usepackage{amssymb}
\usepackage[ruled]{algorithm}
\usepackage{algpseudocode}
\usepackage{ctable}
\usepackage[export]{adjustbox}
\usepackage{import}
\usepackage{pgf}
\usepackage{setspace}
\usepackage{url}
\usepackage{hyperref}
\usepackage{capt-of,etoolbox}
\usepackage{pifont} 
\usepackage{mathtools}
\newcommand{\cmark}{\ding{51}}%
\newcommand{\xmark}{\ding{55}}%

\usepackage{fontawesome} 
\usepackage{multicol}  
\usepackage{balance}

\makeatletter
\let\NAT@parse\undefined
\makeatother
\usepackage[sort&compress,numbers]{natbib}

\usepackage{Packages/sparo_acronyms}
\usepackage{Packages/sparo_math}
\usepackage{Packages/sparo_SIunits}
\usepackage{Packages/sparo_misc}
\usepackage{multirow}
\usepackage{indentfirst}
\usepackage{makecell}

\usepackage{soul,color}
\usepackage{verbatim} 

\usepackage[font=small,skip=4pt]{caption}
\usepackage{float} 

\usepackage{xcolor}
\usepackage[table]{xcolor}

\newcommand{\rl}[1]{{\textcolor{red}{#1}}}

\definecolor{gl}{HTML}{008000}
\newcommand{\gl}[1]{{\textcolor{gl}{#1}}}

\title{\LARGE \bf
    KISS-IMU: Self-supervised Inertial Odometry \\
    with Motion-balanced Learning and Uncertainty-aware Inference
}
\author{Jiwon Choi$^{1}$, Hogyun Kim$^{1}$, Geonmo Yang$^{1}$, Juhui Lee$^{1}$, and Younggun Cho$^{1\dagger}$
        \thanks{
        This work was supported by National Research Foundation of Korea (NRF) grant (RS-2025-02217000) funded by the Korea government (MSIT) \& This work was supported by National Research Foundation of Korea (NRF) grant (RS-2025-24803365) funded by the Korea government (MSIT) \& This work was supported by Institute of Information \& communications Technology Planning \& Evaluation (IITP) grant funded by the Korea government(MSIT) (No.2022-0-00448/RS-2022-II220448) and Inha University.
        }
	\thanks{$^{1}$Jiwon Choi, $^{1}$Hogyun Kim, $^{1}$Geonmo Yang, $^{1}$Juhui Lee, and $^{1\dagger}$Younggun Cho are with the Electrical and Computer Engineering, Inha University, Incheon, South Korea. {\tt\small [jiwon2, hg.kim, ygm7422, dlwngml6635]@inha.edu, yg.cho@inha.ac.kr} \hfill \break 
  }%
}

\newcommand{\gt}{ground truth }

\begin{document}
\maketitle
\begin{abstract}
Inertial measurement units (IMUs), which provide high-frequency linear acceleration and angular velocity measurements, serve as fundamental sensing modalities in robotic systems.
Recent advances in deep neural networks have led to remarkable progress in inertial odometry. 
However, the heavy reliance on ground truth data during training fundamentally limits scalability and generalization to unseen and diverse environments.
We propose \textit{KISS-IMU}, a novel self-supervised inertial odometry framework that eliminates ground truth dependency by leveraging simple LiDAR-based ICP registration and pose graph optimization as a supervisory signal.
Our approach embodies two key principles: keeping the IMU \textit{stable} through motion-aware balanced training and keeping the IMU \textit{strong} through uncertainty-driven adaptive weighting during inference.
To evaluate performance across diverse motion patterns and scenarios, we conducted comprehensive experiments on various real-world platforms, including quadruped robots.
Importantly, we train only the IMU network in a self-supervised manner, with LiDAR serving solely as a lightweight supervisory signal rather than requiring additional learnable processes.
This design enables the framework to ensure robustness without relying on joint multi-modal learning or ground truth supervision.
The supplementary materials are available at \texttt{\url{https://sparolab.github.io/research/kiss_imu}}.
\end{abstract}
\definecolor{myemerald}{rgb}{0.753, 0.898, 0.804}
\definecolor{mylightgreen}{rgb}{0.894, 0.933, 0.745}
\definecolor{myyellow}{rgb}{0.996, 0.972, 0.780}
\newcommand{\firstc}{\cellcolor{myemerald!100}}
\newcommand{\secondc}{\cellcolor{mylightgreen!100}}
\newcommand{\thirdc}{\cellcolor{myyellow!100}}
\newcommand{\oursdc}{\cellcolor{white!100}}
\newcommand{\basedc}{\cellcolor{gray!20}}

\newcommand{\seen}{(\,\raisebox{-0.5ex}{\includegraphics[height=2.2ex]{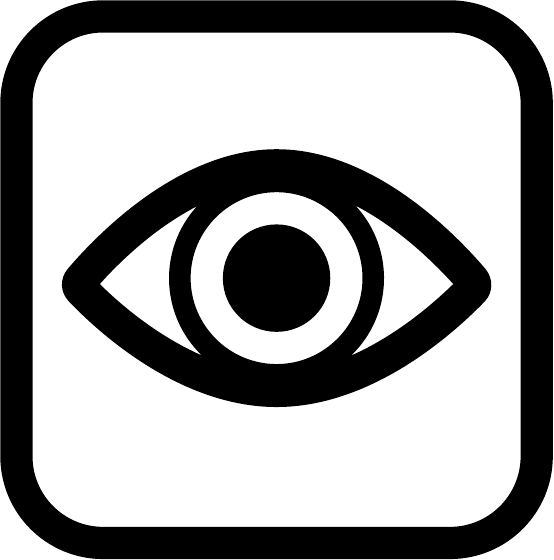}}\,)}
\newcommand{\unseen}{(\,\raisebox{-0.5ex}{\includegraphics[height=2.2ex]{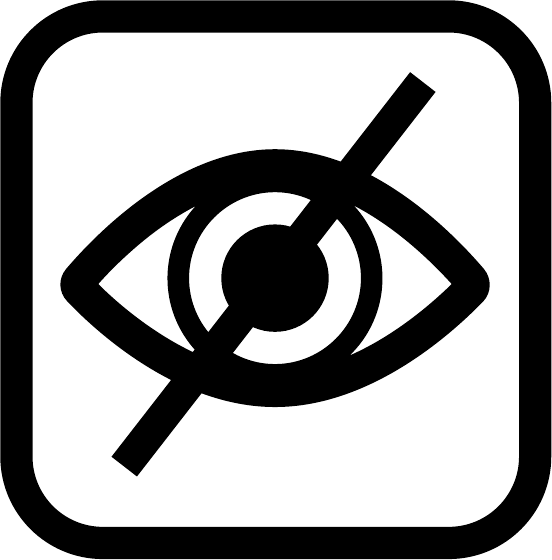}}\,)}

\newcommand{\gmm}{\raisebox{-0.5ex}{\includegraphics[height=2.4ex]{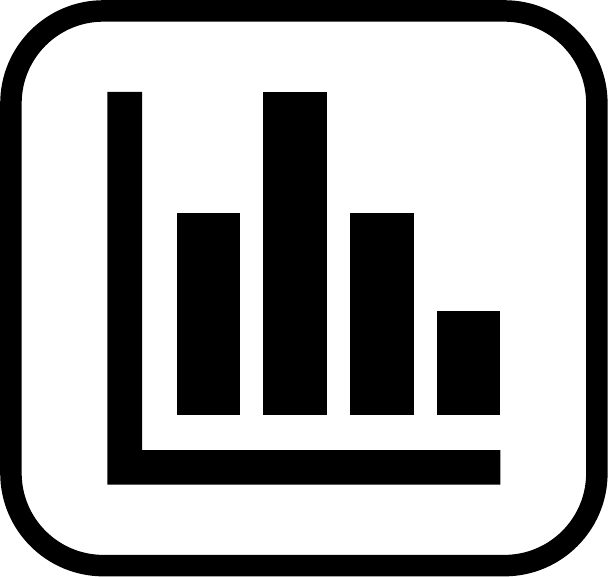}}}
\newcommand{\pgo}{\raisebox{-0.5ex}{\includegraphics[height=2.4ex]{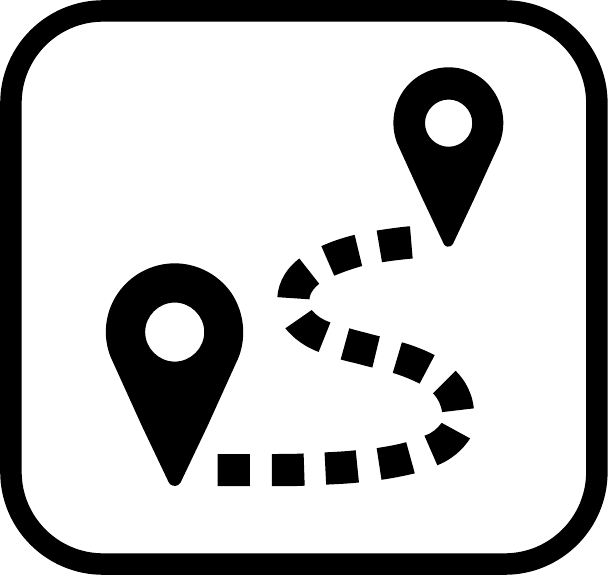}}}
\newcommand{\adw}{\raisebox{-0.5ex}{\includegraphics[height=2.4ex]{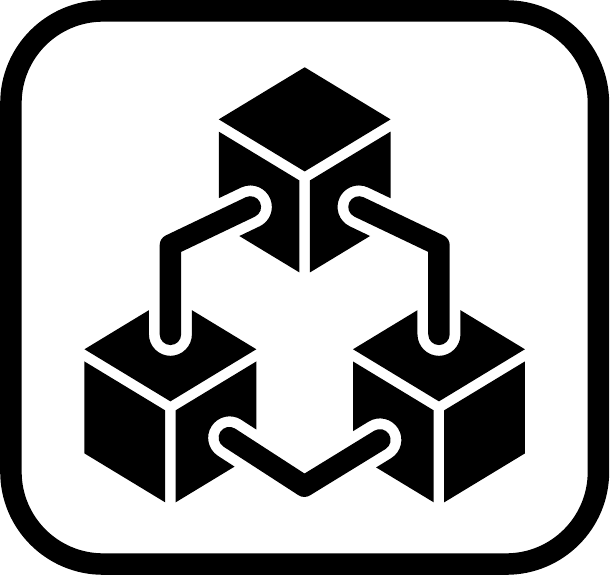}}}

\section{Introduction}

High-frequency measurements of linear acceleration and angular velocity, provided by an inertial measurement unit (IMU), serve as a fundamental basis for analyzing robot motion dynamics and estimation uncertainty. 
Such dense temporal information enables detailed characterization of motion dynamics, which lower-rate exteroceptive sensors fail to capture, providing a reliable basis for understanding and modeling complex robot behaviors.
Consequently, the IMU has become an indispensable sensing modality for understanding and modeling ego-motion across environments  \cite{forster2015imu}.

With recent advances in deep neural networks, inertial odometry (IO) has demonstrated remarkable capabilities by leveraging powerful learning paradigms across various robotic applications~\cite{chen2018ionet, herath2020ronin, liu2020tlio, wang2022llio, brossard2020ai, cioffi2023learned, qiu2023airimu, qiu2025airio, fu2024islam, buchanan2022learning, zhao2025tartan}. 
However, a critical bottleneck remains: the heavy reliance on \gt {data} for training. 
While ground truth provides reliable supervision, obtaining accurate pose data in real-world environments is particularly challenging and resource-intensive.
\begin{figure}[t]
    \centering
    \def\width{0.48\textwidth}%
        {%
     \includegraphics[clip, width=0.48\textwidth]{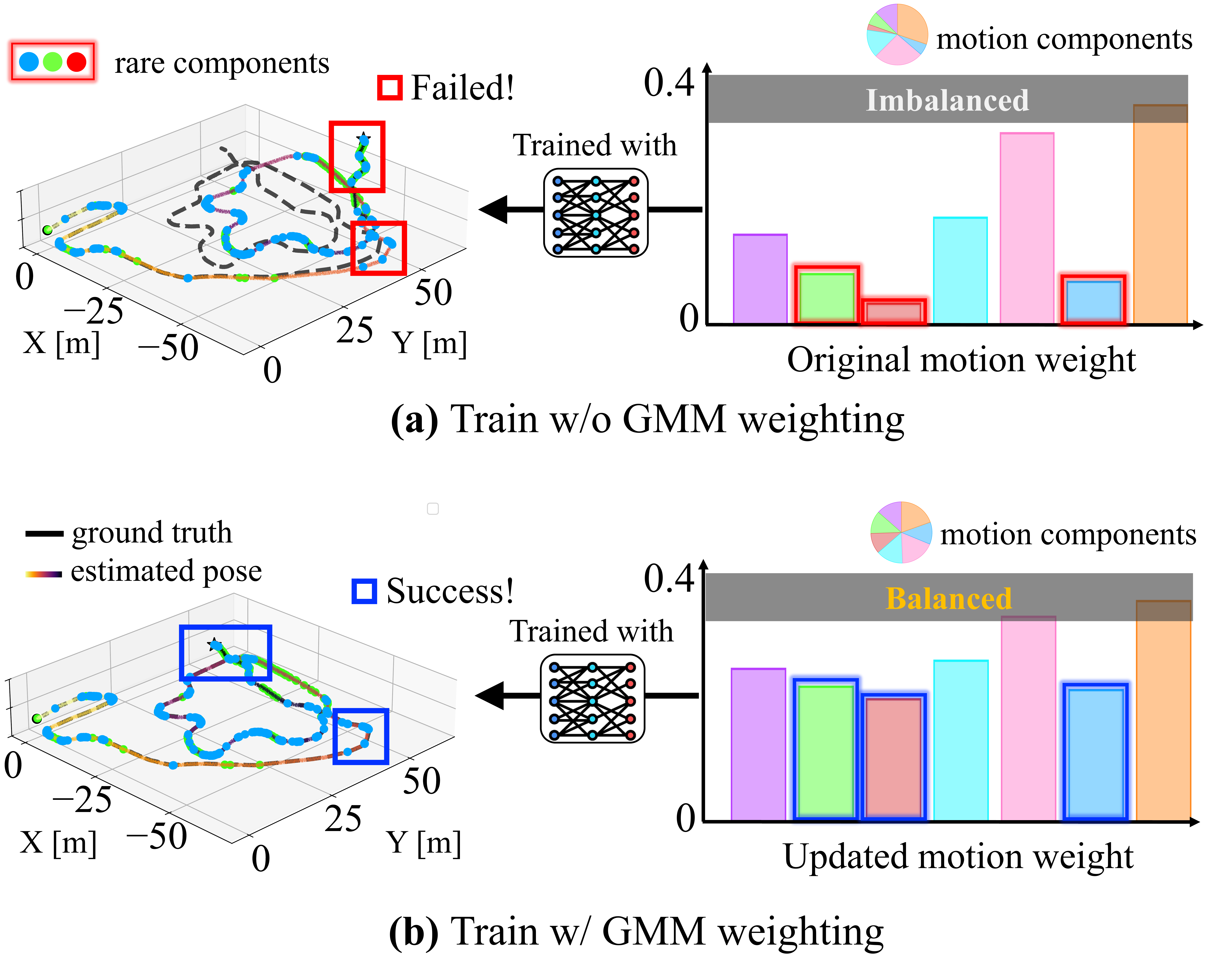}
        }
    \caption{\textit{KISS-IMU} performance on unseen \texttt{LAWN} sequence (trained on \texttt{Forest}, both from \texttt{DiTer++}).
    (a) Training without a Gaussian mixture model (GMM) weighting: Imbalanced motion components bias learning toward dominant patterns, causing trajectory drift and poor generalization. 
    (b) Training with GMM weighting: Our balanced motion components mitigate bias and improve generalization.
    Red boxes in (a) indicate failures in regions with rare motion components, 
    while blue boxes in (b) highlight challenging regions corrected through motion-aware reweighting.
    }
    \vspace{-0.6cm}
    \label{fig:main}
\end{figure}

Recent efforts have attempted to address this limitation through self-supervised learning approaches~\cite{fu2024islam}. 
However, these methods still require joint training across multiple modalities (e.g., visual and inertial), resulting in tightly coupled network architectures that tie performance to specific environmental conditions.
Consequently, this approach inherits the same scalability and generalization limitations as existing supervised methods, leaving these fundamental challenges unresolved.

To overcome these limitations, we propose \emph{KISS-IMU}, a novel \textit{self-supervised} IO framework designed to \emph{'Keep IMU Stable and Strong'}. 
Our main contributions are as follows:
\begin{itemize}
    \item \textbf{Self-supervised Learning Paradigm}: We introduce a novel training approach that eliminates ground truth dependency through selective fusion of LiDAR-based registration and pose graph optimization (PGO) to generate reliable pseudo-labels. This enables scalable deployment in unseen and diverse environments. 

    \item {\textbf{Stable IO through Motion-Aware Training}:} We develop a Gaussian mixture model (GMM)-based motion clustering technique that ensures balanced learning across diverse motion patterns as illustrated in \figref{fig:main}. 
    This approach makes the IMU learning stable by preventing bias toward dominant motions and enhancing comprehensive motion understanding for improved generalization to unseen environments.

    \item {\textbf{Strong IO through Uncertainty-Driven Inference}:} We design an adaptive mechanism that analyzes IMU-based state uncertainty, keeping the IMU state estimation strong by dynamically adjusting weights to maintain robust performance under varying conditions.

    \item {\textbf{Comprehensive Evaluation}:} Through extensive evaluation across multiple datasets under varying training data percentages, we demonstrate competitive generalization capabilities with minimal training data compared to existing state-of-the-art approaches.      
\end{itemize}

To the best of our knowledge, this work introduces a distinct self-supervised paradigm for inertial odometry, where learning is guided by geometry-derived supervision rather than ground truth labels or jointly trained supervisory networks.
 
\section{Related Works} \label{Realted Works}

\subsection{Inertial Odometry For Complex Motions} \label{subsection:imu_odom}
To address the challenges of complex motions in domains such as pedestrians, drones, and quadruped robots, inertial odometry (IO) has been extensively advanced through data-driven methods. 
In pedestrian scenarios, IONet~\cite{chen2018ionet} directly regressed motion using neural networks, and this approach was later improved by RoNIN~\cite{herath2020ronin} through the use of multiple network architectures.
However, both methods suffered from limited interpretability, as measurement noise was not explicitly modeled. 
To overcome this limitation, TLIO~\cite{liu2020tlio} and LLIO~\cite{wang2022llio} predicted uncertainty and integrated it with \ac{EKF} for robust performance.


Unlike pedestrian IO, where periodic gait patterns can be effectively learned, \ac{UAV} domains present a more challenging environment without such regularities.
To better capture real-world dynamics, Drone IO~\cite{cioffi2023learned} predicted displacement and improved measurement updates, rather than depending on assumptions of simplified motion models such as AI-IMU~\cite{brossard2020ai}.
AirIMU~\cite{qiu2023airimu} jointly modeled IMU noise and uncertainty, employing PGO with uncertainty-aware covariance injection.
AirIO~\cite{qiu2025airio} further analyzed the representational richness of motion in the body frame and developed a tightly coupled \ac{EKF} framework.


Similarly, the field expanded to include quadruped robots, which presented unique challenges. 
Despite their promising mobility, their dynamic gaits introduced vibrations across all axes and impact-induced oscillations from ground contact, making motion analysis particularly challenging. 
While approaches such as~\cite{buchanan2022learning} addressed quadruped robot state estimation, these complex dynamics still demanded further investigation.




Despite remarkable progress in learnable IO, existing methods still face a critical limitation: they rely on supervised learning with ground truth data requiring centimeter-level accuracy. 
This dependency necessitates motion capture systems or other high-precision setups. 
Such restricted settings exacerbate the tendency of data-driven methods to overfit specific motion patterns and environments observed during training.
This fundamentally hinders generalization to new scenarios and diverse motions.
iSLAM~\cite{fu2024islam} alleviated this dependency through self-supervised learning, but its reliance on a learnable visual modality left generalization challenges unresolved.

\subsection{IMU Motion Analysis for Enhanced State Estimation}
Sophisticated IMU motion analysis has proven crucial for robust state estimation. 
FAST-LIO2~\cite{xu2022fast} established efficient LiDAR-inertial fusion but suffered from degradation under aggressive motions with IMU saturation. 
Point-LIO~\cite{he2023point} addressed this by treating IMU measurements as signals and conducting separate saturation checks for each channel, enabling accurate localization during extreme motions.
TartanIMU~\cite{zhao2025tartan} showed that learned motion clustering across platforms improves IMU-based state estimation through motion-specific dynamics in feature space.
Super Odometry~\cite{zhao2025resilient} extended this by training inertial odometry across heterogeneous platforms, elevating the IMU to a fallback under sensor degradation.

These works have consistently shown that motion analysis techniques, such as decomposing complex patterns and adapting to motion dynamics, directly impact state estimation performance.
Inspired by these developments, our approach adapts similar motion analysis principles for self-supervised learning scenarios where \gt supervision is unavailable. 
\allowdisplaybreaks  

\section{KISS-IMU: Self-supervised, Stable, and Strong Inertial Odometry} \label{Method}
\begin{figure*}[t]
	\centering
        {%
		\includegraphics[clip, width=0.98\textwidth]{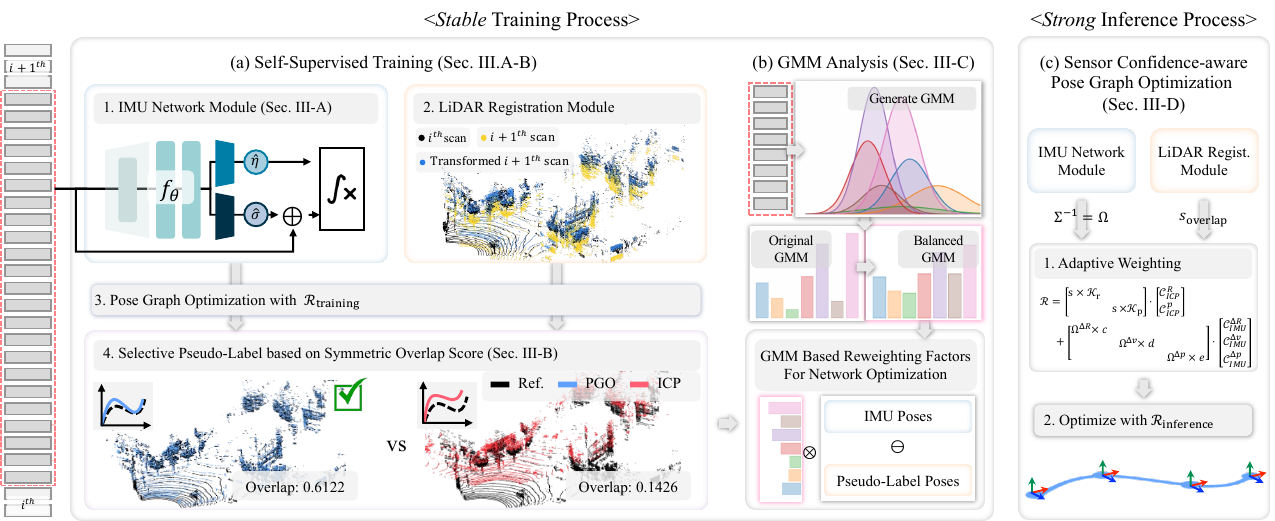}
	  }
        \caption{
        Our inertial odometry (IO) framework follows the \textit{Keep IMU Stable and Strong} philosophy through three components:
        (a) Self-supervised training combines an IMU network for correction and uncertainty prediction with a LiDAR registration module for geometric constraints. 
        Pose graph optimization (PGO) fuses both modalities, followed by selective pseudo-label generation using symmetric overlap scores for supervision without ground truth.
        (b) GMM analysis stabilizes IO via motion clustering. 
        While the original GMM reveals imbalanced motion distributions, our balancing strategy ensures uniform coverage, with reweighting emphasizing underrepresented motions during optimization.
        (c) Sensor confidence-aware PGO strengthens IO through adaptive weighting. 
        Learned uncertainties from IMU and LiDAR modules enable dynamic confidence adjustment during inference, ensuring robustness across varying motion and sensor conditions.
        }
        \vspace{-0.2cm}
\label{fig:pipeline}
\end{figure*}

\begin{figure}[t]
    \centering
    \def\width{0.49\textwidth}%
        {%
\includegraphics[clip, trim={5pt 0 0 0}, width=0.47\textwidth]{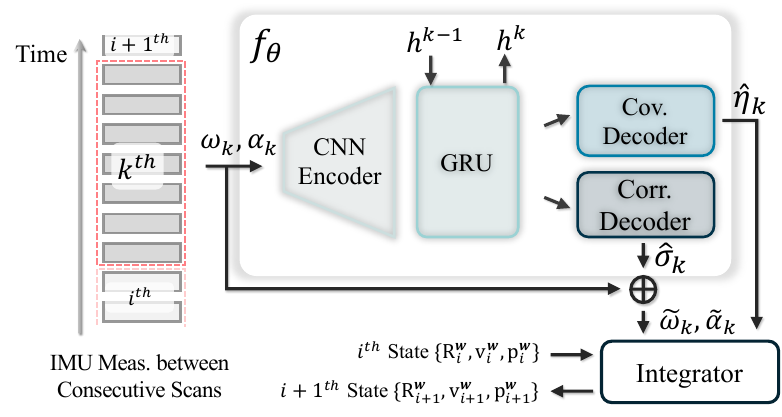}        }
    \caption{
    Our proposed network architecture.
    A CNN-GRU encoder processes raw IMU measurements (i.e., $\mathbf{m}_k = [\boldsymbol{\omega}_k, \boldsymbol{\alpha}_k]^T$) from $\mathcal{M}_{i,i+1}$ to extract features and estimate learned corrections $\hat{\boldsymbol{\sigma}}_k$ and uncertainties $\hat{\boldsymbol{\eta}}_k$.
    An integrator combines these with the previous state to produce corrected measurements in \eqref{equ:corrected_meas} and associated $\hat{\boldsymbol{\eta}}_k$.
    }
    \label{fig:network}
    \vspace{-1cm}
\end{figure}

%
We propose \textit{KISS-IMU}, a \textit{self-supervised} framework for \textit{stable} and \textit{strong} inertial odometry (IO) without ground truth supervision, as illustrated in \figref{fig:pipeline}.
Our method leverages LiDAR registration, \ac{ICP}~\cite{vizzo2023kiss, small_gicp}, and PGO to generate reliable pseudo-labels from segment-wise relative motion ($\Delta\mathbf{R}, \Delta\mathbf{v}, \Delta\mathbf{p}$); see Sec.~\ref{subsec:pseudolabel}.
To achieve \textit{stable} IO, we apply GMM-based reweighting during training to mitigate motion imbalance and prevent bias toward dominant patterns (Sec.~\ref{subsec:loss}).
For \textit{strong} IO, the network jointly estimates IMU corrections and uncertainties, enabling adaptive PGO weighting during inference under varying conditions (Sec.~\ref{subsec:pgo}).

\subsection{IMU Integration and Covariance Propagation} \label{subsec:imu}
Our framework leverages the temporal correspondence between IMU and LiDAR measurements to utilize self-supervised learning. 
For consecutive LiDAR scans between $i^{th}$ time $t_i$ and $i+1^{th}$ time $t_{i+1}$, we aggregate IMU measurements $\mathcal{M}_{i,i+1} = \{\mathbf{m}_k\}_{k=1}^{K}$ within the time interval $[t_i, t_{i+1}]$, where the $k^{th}$ measurement $\mathbf{m}_k = [\boldsymbol{\omega}_k, \boldsymbol{\alpha}_k]^T \in \mathbb{R}^6$ comprises angular velocity, i.e., $\boldsymbol{\omega}_k \in \mathbb{R}^3$, and linear acceleration, i.e., $\boldsymbol{\alpha}_k \in \mathbb{R}^3$.

Unlike learning-free methods that rely on predefined noise models, we utilize a neural network, $f_\theta$ (illustrated in \figref{fig:network}), to adaptively estimate and learn both measurement-specific corrections and uncertainties as follows:
\begin{equation}
    \{\hat{\boldsymbol{\sigma}}_k, \hat{\boldsymbol{\eta}}_k\}_{k=1}^{K} = f_\theta(\mathcal{M}_{i,i+1}),
    \label{equ:learned}
\end{equation}
where $\hat{\boldsymbol{\sigma}}_k = [\hat{\boldsymbol{\sigma}}^{\boldsymbol{\omega}}_k, \hat{\boldsymbol{\sigma}}^{\boldsymbol{\alpha}}_k]^T$ denotes the learned corrections and $\hat{\boldsymbol{\eta}}_k = [\hat{\boldsymbol{\eta}}^{\boldsymbol{\omega}}_k, \hat{\boldsymbol{\eta}}^{\boldsymbol{\alpha}}_k]^T$ denotes the learned uncertainties for angular velocity and acceleration, respectively.

Here, $\hat{\boldsymbol{\sigma}}^{\boldsymbol{\omega}}_k, \hat{\boldsymbol{\sigma}}^{\boldsymbol{\alpha}}_k, \hat{\boldsymbol{\eta}}^{\boldsymbol{\omega}}_k, \hat{\boldsymbol{\eta}}^{\boldsymbol{\alpha}}_k \in \mathbb{R}^3$, resulting in $\hat{\boldsymbol{\sigma}}_k, \hat{\boldsymbol{\eta}}_k \in \mathbb{R}^6$.
Through the learned corrections, we obtain the corrected measurements as follows:
\begin{equation}
    \tilde{\mathbf{m}}_k = \mathbf{m}_k + \hat{\boldsymbol{\sigma}}_k.
\label{equ:corrected_meas}
\end{equation}

Following the preintegration framework~\cite{forster2015imu}, we perform integration consistent with this temporal correspondence to compute the relative transformation between consecutive frames.
The corrected measurements $\tilde{\mathbf{m}}_k$ are then integrated to obtain the preintegrated quantities—rotation $\Delta\mathbf{R}_{i,i+1}^{\text{IMU}} \in SO(3)$, velocity $\Delta\mathbf{v}_{i,i+1}^{\text{IMU}} \in \mathbb{R}^3$, and position $\Delta\mathbf{p}_{i,i+1}^{\text{IMU}} \in \mathbb{R}^3$ as follows:
\begin{subequations}
\label{equ:preintegrate}
\begin{align}
   \Delta\mathbf{R}_{i,i+1}^{\text{IMU}} &= \prod_{k=1}^{K} \text{Exp}(\tilde{\boldsymbol{\omega}}_k\Delta t), \label{equ:rotation}\\
   \Delta\mathbf{v}_{i,i+1}^{\text{IMU}} &= \sum_{k=1}^{K} \Delta\mathbf{R}_{i,k}^{\text{IMU}}\tilde{\boldsymbol{\alpha}}_k\Delta t, \label{equ:velocity}\\
   \Delta\mathbf{p}_{i,i+1}^{\text{IMU}} &= \sum_{k=1}^{K} \left[\Delta\mathbf{v}_{i,k}^{\text{IMU}}\Delta t + \frac{1}{2}\Delta\mathbf{R}_{i,k}^{\text{IMU}}\tilde{\boldsymbol{\alpha}}_k\Delta t^2\right], \label{equ:position}
\end{align}
\end{subequations}
where $\text{Exp}(\cdot)$ denotes the matrix exponential for rotation, $\Delta t$ is the time step between consecutive IMU measurements, and $\Delta\mathbf{R}_{i,k}^{\text{IMU}}$ and $\Delta\mathbf{v}_{i,k}^{\text{IMU}}$ represent the intermediate rotation and velocity from time $t_i$ to $t_k$.
To propagate the IMU state forward in time, we transform the preintegrated measurements to the world frame at time $t_{i+1}$, which is estimated as follows:
\begin{subequations}
\label{equ:world_pre}
\begin{align}
  \hat{\mathbf{R}}_{i+1}^W &= \mathbf{R}_i^W \Delta\mathbf{R}_{i,i+1}^{\text{IMU}}, \label{equ:world_rotation} \\
  \hat{\mathbf{v}}_{i+1}^W &= \mathbf{v}_i^W + \mathbf{g}^W\Delta t + \mathbf{R}_i^W\Delta\mathbf{v}_{i,i+1}^{\text{IMU}}, \label{equ:world_velocity} \\
  \hat{\mathbf{p}}_{i+1}^W &= \mathbf{p}_i^W + \mathbf{v}_i^W\Delta t + \frac{1}{2}\mathbf{g}^W\Delta t^2 + \mathbf{R}_i^W\Delta\mathbf{p}_{i,i+1}^{\text{IMU}}, \label{equ:world_position}
\end{align}
\end{subequations}
where the superscript $W$ denotes quantities in the world frame, $\mathbf{R}_i^W$, $\mathbf{v}_i^W$, and $\mathbf{p}_i^W$ represent the rotation, velocity, and position at the previous time $t_i$ in the world frame, $\hat{\mathbf{R}}_{i+1}^W$, $\hat{\mathbf{v}}_{i+1}^W$, and $\hat{\mathbf{p}}_{i+1}^W$ are the estimated states at time $t_{i+1}$, $\mathbf{g}^W \in \mathbb{R}^3$ is the gravity vector, and $\Delta t = t_{i+1} - t_i$.

A distinguishing feature of our approach is the propagation of learned uncertainties from \equref{equ:learned} through the integration process. 
Following~\cite{qiu2023airimu}, the state covariance evolves iteratively for each measurement step as follows:
\begin{align} \label{equ:learned_cov}
    \boldsymbol{\Sigma}_{k+1} 
    &= \mathbf{A}_k\boldsymbol{\Sigma}_{k}\mathbf{A}_k^T 
    + \mathbf{B}_{\boldsymbol{\omega},k}\,\text{diag}(\hat{\boldsymbol{\eta}}^{\boldsymbol{\omega}}_k)\,\mathbf{B}_{\boldsymbol{\omega},k}^T \nonumber \\
    &\quad\quad\quad\quad\quad\quad\quad\quad\quad\quad 
    + \mathbf{B}_{\boldsymbol\alpha,k}\,\text{diag}(\hat{\boldsymbol{\eta}}^{\boldsymbol\alpha}_k)\,\mathbf{B}_{\boldsymbol{\alpha},k}^T,
\end{align}
where $\mathbf{A}_k$ is the state transition matrix and $\mathbf{B}_{\boldsymbol{\omega},k}$, $\mathbf{B}_{\boldsymbol{\alpha},k}$ are noise propagation matrices that map measurement uncertainties to the state space.
These matrices follow the standard IMU error-state formulation~\cite{forster2015imu, qiu2023airimu}.
The covariance propagation starts with initial covariance $\boldsymbol{\Sigma}_0$ and iteratively accumulates the learned uncertainties $\hat{\boldsymbol{\eta}}^{\boldsymbol\omega}_k$ and $\hat{\boldsymbol{\eta}}^{\boldsymbol\alpha}_k$ for each corrected measurement $\tilde{\mathbf{m}}_k$ from~\eqref{equ:corrected_meas}.
After processing all $k \in \{1, \dots, K\}$ within $[t_i, t_{i+1}]$, the corrected propagation covariance is $\boldsymbol{\Sigma}_{i,i+1} = \boldsymbol{\Sigma}_K$.

This learned and corrected uncertainty $\boldsymbol{\Sigma}_{i,i+1}$ plays two critical roles in our framework. 
During training, it enables uncertainty-aware learning (for \textit{stable} IO in Section~\ref{subsec:loss}). 
During inference, it determines the confidence of IMU constraints in PGO (for \textit{strong} IO in Section~\ref{subsec:pgo}).

\subsection{Selective Pseudo-Labels for Self-Supervised Learning} \label{subsec:pseudolabel}

Our self-supervised framework removes ground truth dependency by generating pseudo-labels from ICP or its variants~\cite{vizzo2023kiss, small_gicp}, 
with geometric-consistency-based selection between ICP and PGO for stable training.
First, we obtain the relative transformation $\Delta\mathbf{T}^{\text{ICP}}_{i,i+1} \in \text{SE}(3)$ between consecutive LiDAR scans at times $t_i$ and $t_{i+1}$.
While ICP provides robust constraints, we enhance robustness by addressing (i) degradation in geometrically degenerate environments and (ii) local optimization limitations affecting global trajectory consistency. 
To obtain more reliable pseudo-labels, we formulate a PGO that jointly considers LiDAR and IMU constraints over $N_{\text{node}}$ frames as follows:
\begin{subequations}
\label{equ:pgo_costs}
\begin{align}
\mathcal C_{\text{ICP}}^{\Delta\mathbf T}
&\coloneqq
\sum_{i=1}^{N_{\text{node}}-1}
\left\| \mathrm{Log}\!\left(\Delta\mathbf T_{i,i+1}^{\mathrm{ICP}} \boxminus \Delta\mathbf T_{i,i+1}^{\mathrm{PGO}} \right)
\right\|^2_{\boldsymbol{\Omega}_{\text{ICP}}^{\Delta\mathbf T}},
\label{equ:cost_icp} \\[4pt]
\mathcal C_{\text{IMU}}^{\Delta\mathbf R}
&\coloneqq
\sum_{i=1}^{N_{\text{node}}-1}
\left\|
\mathrm{Log}\!\left(
\Delta\mathbf R_{i,i+1}^{\mathrm{IMU}}
\boxminus
\Delta\mathbf R_{i,i+1}^{\mathrm{PGO}}
\right)
\right\|^2_{\boldsymbol{\Omega}_{\text{IMU}}^{\Delta\mathbf R}},
\label{equ:cost_imu_rotation} \\[4pt]
\mathcal C_{\text{IMU}}^{\Delta\mathbf v}
&\coloneqq
\sum_{i=1}^{N_{\text{node}}-1}
\left\|
\Delta\mathbf v_{i,i+1}^{\mathrm{IMU}}
-
\Delta\mathbf v_{i,i+1}^{\mathrm{PGO}}
\right\|^2_{\boldsymbol{\Omega}_{\text{IMU}}^{\Delta\mathbf v}},
\label{equ:cost_imu_velocity} \\[4pt]
\mathcal C_{\text{IMU}}^{\Delta\mathbf p}
&\coloneqq
\sum_{i=1}^{N_{\text{node}}-1}
\left\|
\Delta\mathbf p_{i,i+1}^{\mathrm{IMU}}
-
\Delta\mathbf p_{i,i+1}^{\mathrm{PGO}}
\right\|^2_{\boldsymbol{\Omega}_{\text{IMU}}^{\Delta\mathbf p}},
\label{equ:cost_imu_position}
\end{align}
\end{subequations}

\noindent
where $\|\cdot\|^2_{\boldsymbol{\Omega}}$ denotes the Mahalanobis distance,
$\Delta\mathbf{T}^{\text{ICP}}$, $\Delta\mathbf{T}^{\text{PGO}} \in \text{SE}(3)$ are relative poses from ICP and PGO,
$\Delta\mathbf{R}^{\text{IMU}}$, $\Delta\mathbf{v}^{\text{IMU}}$, $\Delta\mathbf{p}^{\text{IMU}}$ are preintegrated IMU quantities from \equref{equ:preintegrate},
and $\boxminus$, $\mathrm{Log}(\cdot)$ denote standard manifold operations.
Each information matrix $\boldsymbol{\Omega}$ is a fixed diagonal matrix constructed from a scalar weight:
$\boldsymbol{\Omega}_{\text{ICP}}^{\Delta\mathbf T} = w_1 \mathbf{I}_6$,
$\boldsymbol{\Omega}_{\text{IMU}}^{\Delta\mathbf R} = w_2 \mathbf{I}_3$,
$\boldsymbol{\Omega}_{\text{IMU}}^{\Delta\mathbf v} = w_3 \mathbf{I}_3$, and
$\boldsymbol{\Omega}_{\text{IMU}}^{\Delta\mathbf p} = w_4 \mathbf{I}_3$,
where $\mathbf{I}_n$ denotes the $n{\times}n$ identity matrix, remaining constant throughout training.
The PGO optimization objective used during training is then:
\begin{equation}
\mathcal{C}_{\text{training}}
\coloneqq
\mathcal C_{\text{ICP}}^{\Delta\mathbf T}
+
\mathcal C_{\text{IMU}}^{\Delta\mathbf R}
+
\mathcal C_{\text{IMU}}^{\Delta\mathbf v}
+
\mathcal C_{\text{IMU}}^{\Delta\mathbf p},
\label{equ:C_pgo}
\end{equation}
\noindent
which is minimized using the Levenberg--Marquardt algorithm in PyPose~\cite{wang2023pypose}.

When conditions (i) and (ii) are not met, ICP provides sufficiently reliable pseudo-labels.
To ensure optimal reliability and select the most trustworthy measurements across all scenarios, we evaluate both ICP and PGO estimates based on their geometric consistency.
For this selective fusion, we assess the quality of each estimate using the symmetric overlap score from~\cite{jung2025helios} as follows:
\begin{equation}
    s_{\text{overlap}}(\Delta\mathbf{T}) = 0.5\left[\mathbf{O}(\mathcal{P}_i, {\mathcal{P}}'_{i+1}) + \mathbf{O}(\mathcal{P}_{i+1}, {\mathcal{P}}'_i)\right],
\label{equ:overlap}
\end{equation}
\noindent where $\mathbf{O}(\cdot, \cdot)$ measures the overlap ratio based on nearest neighbor distances, and ${\mathcal{P}}'_{i+1}$, ${\mathcal{P}}'_i$ represent the transformed point clouds from $\Delta\mathbf{T}$.
We evaluate both $s_{\text{ICP}} = s_{\text{overlap}}(\Delta\mathbf{T}_{i,i+1}^{\text{ICP}})$ and $s_{\text{PGO}} = s_{\text{overlap}}(\Delta\mathbf{T}_{i,i+1}^{\text{PGO}})$, selecting the transformation with the higher overlap score as our pseudo-label.
\figref{fig:pipeline}\,(a).\,4 demonstrates this selection process, where the higher overlap score indicates better geometric alignment between consecutive point clouds, providing a more trustworthy reference pose for supervision.
Moreover, this selective mechanism helps prevent the network from reinforcing its own errors by ensuring supervision comes from the most geometrically consistent source.

We denote the selected pseudo-label as $\Delta\mathbf{T}_{i,i+1} = \{\Delta\mathbf{R}_{i,i+1}^{\text{pseudo}}, \Delta\mathbf{p}_{i,i+1}^{\text{pseudo}}\}$, representing the relative transformation between consecutive frames.
To derive the supervisory states for training, we propagate the pose: given $\mathbf{T}_i^W = \{\mathbf{R}_i^W, \mathbf{p}_i^W\}$ at time $t_i$, the pose at $t_{i+1}$ is computed as $\mathbf{T}^{W}_{i+1} = \mathbf{T}^{W}_i \cdot \Delta\mathbf{T}_{i,i+1}$, and the world-frame velocity as $\mathbf{v}_{i+1}^W = (\mathbf{p}_{i+1}^W - \mathbf{p}_i^W) /\Delta t$, where $\Delta t$ is defined in \eqref{equ:world_pre}.
These pseudo-label states $\{\mathbf{R}_{i+1}^W, \mathbf{v}_{i+1}^W, \mathbf{p}_{i+1}^W\}$ serve as supervision for the IMU-predicted states in our loss functions (Section~\ref{subsec:loss}), enabling the network to learn accurate IO without ground~truth.

\subsection{Loss Functions and GMM-Based Training Strategy} \label{subsec:loss}
\subsubsection{Loss Functions}
We first define the pose-level loss functions for rotation, velocity, and position errors between the predicted states $\{\hat{\mathbf{R}}_{i+1}^W, \hat{\mathbf{v}}_{i+1}^W, \hat{\mathbf{p}}_{i+1}^W\}$ and pseudo-labels $\{\mathbf{R}_{i+1}^W, \mathbf{v}_{i+1}^W, \mathbf{p}_{i+1}^W\}$ as follows:
\begin{subequations}
\begin{align}
   \mathcal{L}_r &= \left\|\log (\mathbf{R}^W_{i+1} \boxminus \hat{\mathbf{R}}_{i+1}^W) \right\|_2, \label{equ:loss_rotation} \\
   \mathcal{L}_v &= \left\| \mathbf{v}^W_{i+1} - \hat{\mathbf{v}}_{i+1}^W \right\|_2, \label{equ:loss_velocity} \\
   \mathcal{L}_p &= \left\| \mathbf{p}^W_{i+1} - \hat{\mathbf{p}}_{i+1}^W \right\|_2. \label{equ:loss_position}
\end{align}
\end{subequations}

\noindent Since each pose estimation inherently contains uncertainty, we additionally define uncertainty-aware loss functions that incorporate the learned uncertainties, derived from \eqref{equ:learned_cov}, to handle prediction reliability better, following the formulations in \cite{qiu2023airimu, russell2021multivariate}:
\begin{subequations}
\begin{align}
   \mathcal{L}_r^{\text{cov}} &= \frac{1}{2}\left(\left\|\log (\mathbf{R}^W_{i+1} \boxminus \hat{\mathbf{R}}_{i+1}^W) \right\|_{\boldsymbol{\Sigma}_{i,i+1}^r}^2 + \ln(\det(\boldsymbol{\Sigma}_{i,i+1}^r))\right), \label{equ:loss_rotation_cov} \\
   \mathcal{L}_v^{\text{cov}} &= \frac{1}{2}\left(\left\| \mathbf{v}^W_{i+1} - \hat{\mathbf{v}}_{i+1}^W \right\|_{\boldsymbol{\Sigma}_{i,i+1}^v}^2 + \ln(\det(\boldsymbol{\Sigma}_{i,i+1}^v))\right), \label{equ:loss_velocity_cov} \\
   \mathcal{L}_p^{\text{cov}} &= \frac{1}{2}\left(\left\| \mathbf{p}^W_{i+1} - \hat{\mathbf{p}}_{i+1}^W \right\|_{\boldsymbol{\Sigma}_{i,i+1}^p}^2 + \ln(\det(\boldsymbol{\Sigma}_{i,i+1}^p))\right), \label{equ:loss_position_cov}
\end{align}
\end{subequations}
where $\boldsymbol{\Sigma}_{i,i+1}^r$, $\boldsymbol{\Sigma}_{i,i+1}^v$, and $\boldsymbol{\Sigma}_{i,i+1}^p$ are the rotation, velocity, and position blocks of the propagated covariance $\boldsymbol{\Sigma}_{i,i+1}$ from \eqref{equ:learned_cov}.
Then, the final loss function combines all components as follows:

\begin{equation}
    \mathcal{L}_{\text{total}} = \underbrace{\mathcal{L}_r + \mathcal{L}_v + \mathcal{L}_p}_{\text{pose-level loss}} + \underbrace{\varepsilon( \mathcal{L}_r^{\text{cov}} + \mathcal{L}_v^{\text{cov}} + \mathcal{L}_p^{\text{cov}})}_{\text{uncertainty-aware loss}},
\label{equ:air_loss}
\end{equation}
where $\varepsilon$ is a scaling factor.

This comprehensive loss formulation allows the network to simultaneously learn accurate pose estimation and reliable uncertainty quantification, yield robust state representation that adapts to varying motion conditions and measurement~quality.

\subsubsection{GMM-Based Motion Decomposition}
While function \eqref{equ:air_loss} effectively incorporates uncertainty, they alone cannot resolve the fundamental challenge of motion distribution imbalance in sequential datasets. 
Such datasets are typically dominated by common motions (e.g., straight-line motion in driving datasets), while rare but critical maneuvers (e.g., sharp yaw or sudden acceleration) are underrepresented. Therefore, we aim to achieve robust IMU learning that balances diverse motion patterns during training, preventing overfitting to frequent behaviors while preserving performance on rare yet essential maneuvers, as shown in \figref{fig:pipeline}(b).

To achieve comprehensive motion coverage, we first define IMU windows of duration $\Delta t_w$ (typically 0.2\,sec) and extract motion descriptors for each window. 
For each window, we extract motion features including statistics of angular velocity, magnitudes of acceleration, and their temporal variations. 
These features are standardized to form a motion descriptor $\mathbf{z}_n \in \mathbb{R}^n$. 
We then analyze the motion distribution of the entire training dataset using these descriptors with GMM similar to \cite{zhao2025tartan}.
We fit a GMM to the collection of motion descriptors from all training sequences, $ p(\mathbf{z}_n) = \sum_{g=1}^{G} \pi_g \mathcal{N}(\mathbf{z}_n \mid \boldsymbol{\mu}_g, \boldsymbol{\Sigma}_g)$, where $G$ is the number of components, and $\pi_g$, $\boldsymbol{\mu}_g$, $\boldsymbol{\Sigma}_g$ represent the mixture weights, means, and covariances respectively. 
Following~\cite{wan2019novel}, we select the optimal number of components $G$ using the Bayesian information criterion (BIC) to balance model complexity and fit quality.

\subsubsection{Motion-Aware Reweighting Strategy}
Inspired by the class-balanced weighting strategy~\cite{cui2019class}, we extend this concept to address motion imbalance in IMU data. 
While they tackle long-tailed distributions in visual recognition, we adapt their reweighting principle to balance diverse motion patterns in~IO.

Let $w_g = \frac{1-\beta}{1-\beta^{N_g}}$ denote the raw reweighting factor for component $g$, where $\beta \in (0,1)$ controls the re-weighting strength. 
Here, $N_g = \sum_{g=1}^{N} \gamma_g(\mathbf{z}_n)$ represents the frequency of each component across the training set, and $\gamma_g(\mathbf{z}_n) = \frac{\pi_g \mathcal{N}(\mathbf{z}_n \mid \boldsymbol{\mu}_g, \boldsymbol{\Sigma}_g)}{\sum_{g=1}^{G} \pi_g \mathcal{N}(\mathbf{z}_n \mid \boldsymbol{\mu}_g, \boldsymbol{\Sigma}_g)}$ denotes the posterior probability of component assignment. 
The normalized weight is then computed as $\tilde{w}_g = \frac{w_g}{\frac{1}{G}\sum_{g=1}^{G}w_g}$. 
Components with lower frequency $N_g$ receive higher weights $\tilde{w}_g$, ensuring rare motion patterns are adequately represented during training.

\begin{figure}[t]
    \centering
    \def\width{0.5\textwidth}%
        {%
     \includegraphics[clip, width=0.48\textwidth]{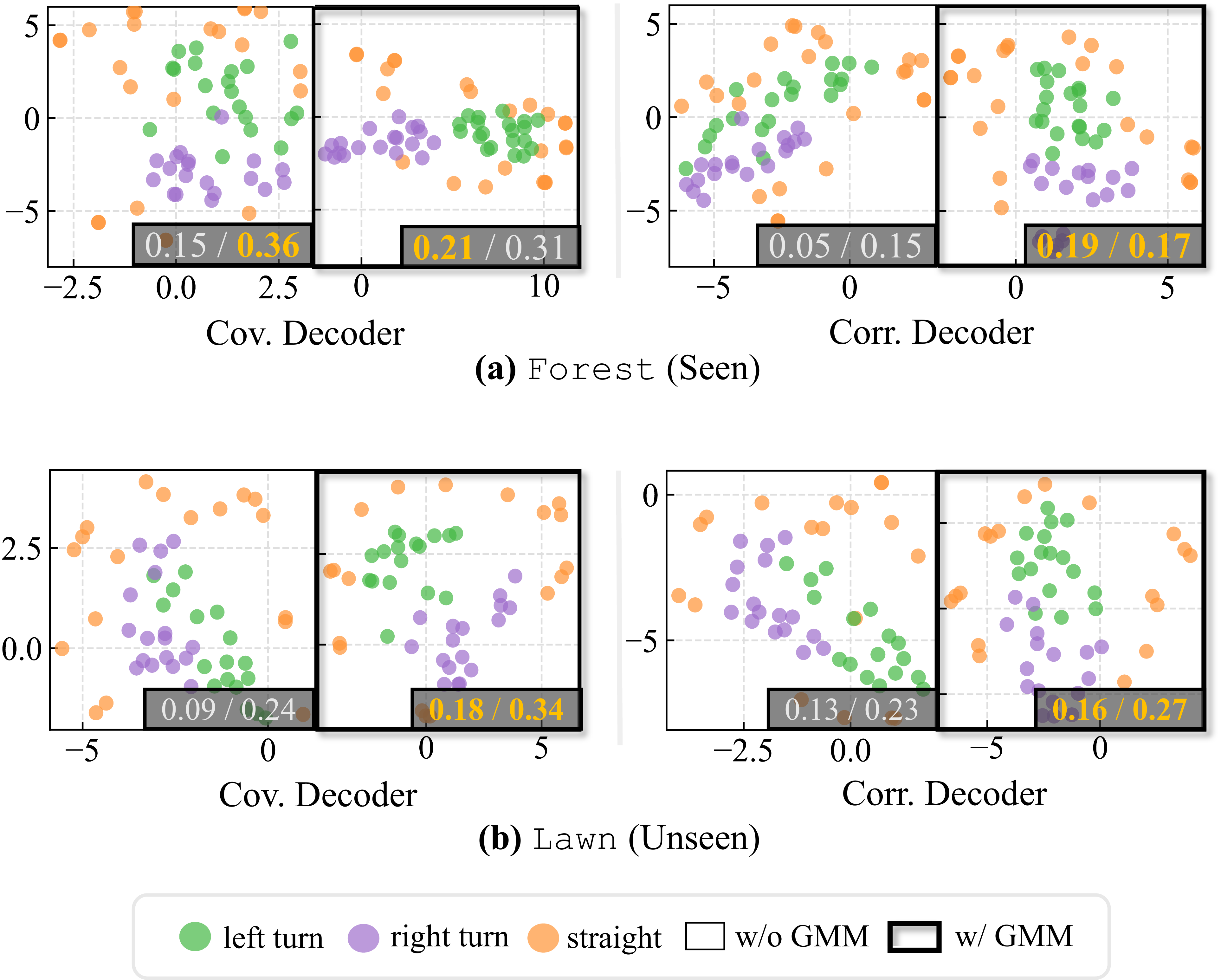}
        }
    \caption{
    t-SNE visualization of motion pattern clustering with quantitative evaluation using silhouette score~\cite{rousseeuw1987silhouettes} and adjusted rand index~\cite{hubert1985comparing}. 
    Motion samples from GMM-determined labels (left turn, right turn, straight), identified via ground truth velocity analysis, are shown in seen (\texttt{Forest}) and unseen (\texttt{LAWN}) environments.
    Without GMM balancing, clustering exhibits limited separation, whereas GMM-based training improves inter-class separation and intra-class compactness.
    These results indicate better generalization in embedding representations through motion-aware reweighting.
    }
\label{fig:tsne}
\vspace{-0.6cm}
\end{figure}
\begin{figure*}[t]
    \centering
    \def\width{0.98\textwidth}%
        {%
     \includegraphics[clip, width=0.98\textwidth]{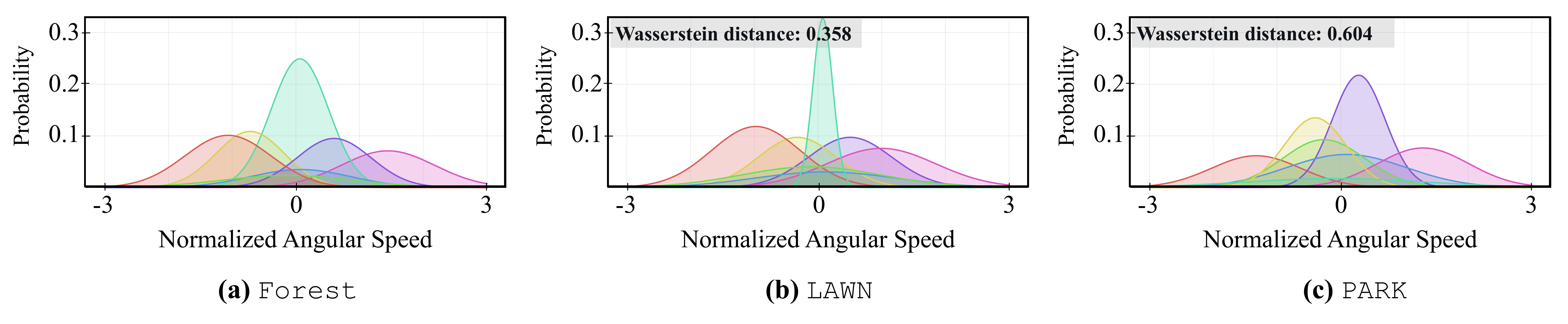}
        }
    \vspace{-0.2cm}
    \caption{Motion pattern analysis using GMM across \texttt{DiTer++} sequences. 
             Using Bayesian information criterion \cite{wan2019novel}, we determine the optimal number of components as $G$=7. 
             GMM distributions of normalized angular speed reveal motion characteristics for (a) \texttt{Forest}, (b) \texttt{LAWN}, and (c) \texttt{PARK} sequences. 
             Lower Wasserstein distances between \texttt{Forest} and \texttt{LAWN} compared to \texttt{PARK} indicate better motion similarity.}
    \label{fig:gmm}
    \vspace{-0.5cm}
\end{figure*}

During training, for each IMU window with motion descriptor $\mathbf{z}_n$, we compute its motion-aware weight as follows:
\begin{equation}
    w_n^{\text{GMM}} = \sum_{g=1}^{G} \gamma_g(\mathbf{z}_n) \tilde{w}_g.
\end{equation}
This weight reflects how much each sample should contribute to the loss based on its motion component assignment and the rarity of that component.
Extending \eqref{equ:air_loss}, the final training loss integrates these motion-aware weights as follows:

\begin{equation}
\mathcal{L}_{\text{total}}^\star = \frac{1}{|\mathcal{B}|} \sum_{n \in \mathcal{B}} w_n^{\text{GMM}} \cdot \mathcal{L}_{\text{total}},
\end{equation}
where $\mathcal{B}$ is the mini-batch and $|\cdot|$ denotes the cardinality of the set.
The per-sample loss combines both motion-aware pose-level and motion- and uncertainty-aware terms as follows:
\begin{align}
   \mathcal{L}_{\text{total}}^\star &= \frac{1}{|\mathcal{B}|} \sum_{n \in \mathcal{B}} \bigg(\underbrace{\lambda_r \mathcal{L}_{r,n} + \lambda_v \mathcal{L}_{v,n} + \lambda_p \mathcal{L}_{p,n}}_{\text{motion-aware pose-level loss}} \nonumber\\
   &\quad\quad\quad\quad\quad\quad + \underbrace{\lambda_r \mathcal{L}_{r,n}^{\text{cov}} + \lambda_v \mathcal{L}_{v,n}^{\text{cov}} + \lambda_p \mathcal{L}_{p,n}^{\text{cov}}}_{\text{motion- and uncertainty-aware loss}}\bigg),
\end{align}
where $\lambda_{r,v,p} = w_n^{\text{GMM}}$ for motion-aware standard terms and $\lambda_{r,v,p} = \varepsilon \cdot w_n^{\text{GMM}}$ for motion- and uncertainty-aware terms.

This motion-aware training strategy ensures balanced learning across diverse motion patterns, emphasizing motion diversity over quantity to prevent the network from overfitting to dominant behaviors while neglecting critical but rare maneuvers. 
By integrating motion-balanced weighting with uncertainty-aware losses, our approach achieves stable IO through comprehensive motion coverage rather than simply increasing training data volume.
As illustrated in \figref{fig:tsne}, our strategy demonstrates improved feature clustering and separation compared to the baseline, indicating enhanced generalization performance across diverse motion patterns.

\subsection{Sensor Confidence-Aware Adaptive Weights for PGO}\label{subsec:pgo}
To achieve strong IO during inference, we enhance PGO through adaptive constraint weighting based on sensor confidence. 
While training employs fixed weights for strong IO, inference benefits from dynamic weighting that adapts to the varying reliability levels of measurements, maintaining robust performance across different motion scenarios.

For adaptive LiDAR constraints, the symmetric overlap score $s_{\text{overlap}}$ from \eqref{equ:overlap} inherently quantifies registration confidence—higher overlap indicates more reliable geometric alignment. 
For adaptive IMU constraints, the propagated covariance $\boldsymbol{\Sigma}_{i,i+1}$ from \eqref{equ:learned_cov} represents measurement uncertainty, which we convert to confidence through the information matrix $\boldsymbol{\Omega}_{i,i+1} = \boldsymbol{\Sigma}_{i,i+1}^{-1}$.
We incorporate these confidence measures into the PGO framework. 
In sum, the adaptive PGO cost function becomes as follows:
\begin{equation}
\begin{aligned}
   \mathcal{C}_{\text{inference}} &= s_{\text{overlap}} \cdot (\kappa_r \|\mathcal{C}_{\text{ICP}}^{\Delta \mathbf{R}}\|^2 + \kappa_p \|\mathcal{C}_{\text{ICP}}^{\Delta\mathbf{p}}\|^2) \\
   &+ \tau_R \|\mathcal{C}_{\text{IMU}}^{\Delta \mathbf{R}}\|_{\boldsymbol{\Omega}^{\Delta\mathbf{R}}}^2 + \tau_v \|\mathcal{C}_{\text{IMU}}^{\Delta\mathbf{v}}\|_{\boldsymbol{\Omega}^{\Delta\mathbf{v}}}^2 + \tau_p \|\mathcal{C}_{\text{IMU}}^{\Delta\mathbf{p}}\|_{\boldsymbol{\Omega}^{\Delta\mathbf{p}}}^2,
\end{aligned}
\end{equation}
where $\kappa_r, \kappa_p$ are LiDAR sensor's scaling factors and $\tau_R, \tau_v, \tau_p$ are IMU sensor's scaling factors.
Finally, by solving this adaptive cost function with the LM optimizer, we achieve strong IO that dynamically adapts to varying sensor reliability and motion conditions.
\section{Experimental Results} \label{Benchmarks}
\subsection{Experimental Setup}
We conducted experiments on a desktop equipped with an AMD EPYC 7513 32-core processor (3.3 GHz) and an NVIDIA RTX 3090 GPU. 
All experiments were performed using PyTorch with CUDA acceleration for neural network training and inference.

\subsubsection{Datasets} We evaluated our method on \texttt{Botanic Garden}~\cite{liu2024botanicgarden}, \texttt{DiTer++}~\cite{kim2025diter++}, and an \texttt{In-House} dataset. 
\texttt{Botanic Garden} provided natural environments captured with a wheeled robot equipped with Velodyne VLP-16 LiDAR and Xsens MTI-680G IMU (9\,DoF), which presented primarily environmental challenges.
\texttt{DiTer++} featured a Unitree GO2 quadrupedal robot with diverse motion patterns across various terrains captured using Ouster OS-1 64\,/\,128 LiDAR and sensor-integrated IMU (6\,DoF), which combined moderate environmental and motion complexities. 
Our \texttt{In-House} dataset used a larger Unitree B2 quadrupedal robot in crater-filled planetary-analogous terrain captured with Ouster OS-1 32 LiDAR and sensor-integrated IMU (6\,DoF), exhibiting extreme motion variations and harsh environmental conditions representing the highest complexity level.
The complexity of each dataset was denoted by stars ($\bigstar$) in our evaluation tables.

\begin{table*}[t]
\captionsetup{width=\textwidth, justification=justified} 
\caption{Quantitative comparison on \texttt{Botanic Garden} and \texttt{DiTer++} datasets with varying training data percentages. 
         Training is performed only on seen sequences ($\raisebox{-0.5ex}{\includegraphics[height=2.2ex]{Figure/icon/seen.pdf}}$: \texttt{1005-01}, \texttt{Forest}), while evaluation includes both seen and unseen sequences ($\raisebox{-0.5ex}{\includegraphics[height=2.2ex]{Figure/icon/unseen.pdf}}$: \texttt{1006-01}, \texttt{1008-03}, \texttt{LAWN}, \texttt{PARK}). 
         Results show that our approach consistently achieves stable performance across different data scales and generalizes better to unseen environments.
         See \tabref{tab:inhouse} footer for details of Ours$^\ddagger$, Ours$^\dagger$, and Ours.}
\centering\resizebox{\textwidth}{!}{\tiny
\begin{tabular}{l|l|cccccccccccc}
\toprule
\hline
\multirow{5}{*}{\hspace{4pt}\rotatebox[origin=c]{90}{Train}}
                  & Dataset         & \multicolumn{6}{c}{\texttt{Botanic Garden} ($\bigstar \bigstar$)}
                                    & \multicolumn{6}{c}{\texttt{DiTer++} ($\bigstar \bigstar \bigstar$)}  \\ \cmidrule(lr){3-8} \cmidrule(lr){9-14}
                  & Seq.            & \multicolumn{2}{c}{\texttt{1005-01}$\seen$}
                                    & \multicolumn{2}{c}{\texttt{1006-01}$\unseen$}
                                    & \multicolumn{2}{c}{\texttt{1008-03}$\unseen$}
                                    & \multicolumn{2}{c}{\texttt{Forest}$\seen$}
                                    & \multicolumn{2}{c}{\texttt{LAWN}$\unseen$}
                                    & \multicolumn{2}{c}{\texttt{PARK}$\unseen$}            \\  \cmidrule(lr){3-4}   \cmidrule(lr){5-6} 
                                                                                                 \cmidrule(lr){7-8}   \cmidrule(lr){9-10} 
                                                                                                 \cmidrule(lr){11-12} \cmidrule(lr){13-14}
                  & Eval.           & RPE\,[m]    & APE\,[m]  
                                    & RPE\,[m]    & APE\,[m]  
                                    & RPE\,[m]    & APE\,[m]
                                    & RPE\,[m]    & APE\,[m]  
                                    & RPE\,[m]    & APE\,[m]  
                                    & RPE\,[m]    & APE\,[m]                     \\ \cline{2-14}

                  & \oursdc Baseline$^{*}$ & \basedc 0.720   & \basedc 15.907  & \basedc 1.180   & \basedc 30.454
                                           & \basedc 0.848   & \basedc 34.450  & \basedc 1.095   & \basedc 6.102   
                                           & \basedc 0.817   & \basedc 9.049   & \basedc 1.160   & \basedc 76.380        \\ \hline 

\multirow{6}{*}{\hspace{4pt}\rotatebox[origin=c]{90}{100\,\%}}
                  & TLIO           & \firstc \textbf{0.372} & 21.164                 & 2.590                  & 66.180   
                                   & 2.956                  & 60.767                 & \firstc \textbf{0.263} & 33.731                         
                                   & 0.700                  & 34.900                 & \firstc \textbf{0.752} & N/A                      \\
                  & AirIO          & \secondc 0.685         & 23.357                 & 1.188                  & 44.769
                                   & \firstc \textbf{0.798} & \secondc 19.443        & 0.916                  & 12.685                 
                                   & 0.652                  & 14.811                 & 1.205                  & \thirdc 70.026          \\
                  & AirIMU$^{*}$   & \secondc 0.685         & 15.884                 & 1.188                  & 41.243
                                   & \firstc \textbf{0.798} & 41.619                 & 0.916                  & 8.099                  
                                   & 0.652                  & 5.014                  & 1.205                  & \secondc 32.982 \\ \cline{2-14}
                  & \oursdc Ours$^\ddagger$ 
                                   & \oursdc 0.712          & \thirdc 4.991          & \firstc \textbf{1.180} & \thirdc 40.306
                                   & \thirdc 0.839          & \oursdc 25.807         & \oursdc 0.880          & \secondc 1.805
                                   & \thirdc 0.644          & \thirdc 1.328          & \oursdc 1.174          & \oursdc 85.252           \\
                  & Ours$^\dagger$ & 0.716                  & \secondc 4.277         & \firstc \textbf{1.180} & \secondc 37.618                 
                                   & \thirdc 0.839          & \thirdc 25.448         & \secondc 0.836         & \thirdc 1.893          
                                   & \firstc \textbf{0.638} & \secondc 1.204         & \secondc 1.159         & 82.396           \\ 
                  & Ours           & 0.716                  & \firstc \textbf{2.531} & \firstc \textbf{1.180} & \firstc \textbf{2.495}                 
                                   & \thirdc 0.839          & \firstc \textbf{7.988} & \secondc 0.836         & \firstc \textbf{1.114}          
                                   & \firstc \textbf{0.638} & \firstc \textbf{0.965} & \secondc 1.159         & \firstc \textbf{9.943}                             \\  \hline \hline

\multirow{6}{*}{\hspace{4pt}\rotatebox[origin=c]{90}{60\,\%}}
                  & TLIO           & 2.507                  & 73.939                 & 2.510                  & 60.374   
                                   & 2.911                  & 67.076                 & \firstc \textbf{0.302} & 9.112                        
                                   & \thirdc 0.681          & 52.309                 & \firstc \textbf{0.693} & N/A                           \\
                  & AirIO          & \firstc \textbf{0.661} & 6.531                  & 1.179                  & \secondc 12.450
                                   & \firstc \textbf{0.796} & 44.696                 & 0.922                  & 33.523                 
                                   & 0.683                  & 29.224                 & 1.208                  & 116.882                       \\
                  & AirIMU$^{*}$   & \firstc \textbf{0.661} & 14.369                 & 1.179                  & \thirdc 29.044
                                   & \firstc \textbf{0.796} & 43.023                 & 0.922                  & 13.047                 
                                   & 0.683                  & 6.478                  & 1.208                  & \secondc 31.905                \\ \cline{2-14}
                  &  \oursdc Ours$^\ddagger$ 
                                   & \oursdc 0.707          & \secondc 5.164         & \thirdc 1.178          & \oursdc 41.046
                                   & \oursdc 0.836          & \secondc 26.080        & \oursdc 0.994          & \firstc \textbf{1.024}
                                   & \oursdc 0.725          & \thirdc 2.408          & \oursdc 1.156          & \thirdc 59.049                  \\
                  & Ours$^\dagger$ & \thirdc 0.702          & \thirdc 5.542          & \firstc \textbf{1.177} & 42.815               
                                   & \thirdc 0.835          & \thirdc 26.291         & \secondc 0.861         & \thirdc 1.716          
                                   & \firstc \textbf{0.632} & \secondc 1.074         & \secondc 1.134         & 80.294                          \\
                  & Ours           & \thirdc 0.702          & \firstc \textbf{2.523} & \firstc \textbf{1.177} & \firstc \textbf{2.712}                 
                                   & \thirdc 0.835          & \firstc \textbf{8.379} & \secondc 0.861         & \secondc 1.052           
                                   & \firstc \textbf{0.632} & \firstc \textbf{0.464} & \secondc 1.134         & \firstc \textbf{6.356}           \\  \hline \hline

\multirow{6}{*}{\hspace{4pt}\rotatebox[origin=c]{90}{20\,\%}}
                  & TLIO           & 2.480                  & 68.257                 & 2.510                  & 58.767   
                                   & 2.917                  & 71.386                 & \firstc \textbf{0.635} & 29.428                         
                                   & \firstc \textbf{0.625} & 45.947                 & \firstc \textbf{0.718} & N/A                                 \\
                  & AirIO          & 0.915                  & 25.834                 & 1.228                  & 45.406
                                   & 1.024                  & 41.396                 & 1.043                  & 16.317                 
                                   & 0.735                  & 26.045                 & 1.347                  & 90.908                              \\
                  & AirIMU$^{*}$   & 0.915                  & 13.746                 & 1.228                  & \secondc 28.960
                                   & 1.024                  & 34.841                 & 1.043                  & 10.775                 
                                   & 0.735                  & 3.526                  & 1.347                  & \secondc 28.269                     \\ \cline{2-14}
                  & \oursdc Ours$^\ddagger$ 
                                   & \thirdc 0.712          & \thirdc 5.326          & \thirdc 1.180          & \thirdc 42.582
                                   & \thirdc 0.860          & \secondc 25.944        & \oursdc 0.997          & \firstc \textbf{0.900}
                                   & \oursdc 0.726          & \secondc 2.033         & \oursdc 1.162          & \thirdc 64.372                       \\
                  & Ours$^\dagger$ & \firstc \textbf{0.695} & \secondc 5.202         & \firstc \textbf{1.178} & 42.688                
                                   & \firstc \textbf{0.828} & \thirdc 26.257         & \secondc 0.945         & \thirdc 1.235          
                                   & \secondc 0.680         & \thirdc 2.074          & \secondc 1.139         & 69.714                         \\               
                  & Ours           & \firstc \textbf{0.695} & \firstc \textbf{2.516} & \firstc \textbf{1.178} & \firstc \textbf{2.716}                 
                                   & \firstc \textbf{0.828} & \firstc \textbf{8.269} & \secondc 0.945         & \secondc 1.059           
                                   & \secondc 0.680         & \firstc \textbf{0.493} & \secondc 1.139         & \firstc \textbf{1.387}                  \\ 
                                   \hline\bottomrule
\end{tabular}}
\label{tab:main}
\vspace{-0.2cm}
\end{table*}

\subsubsection{Evaluation Metrics}
We evaluated our method using two trajectory accuracy metrics: absolute pose error (APE) and relative pose error (RPE).
In particular, to focus on motion errors rather than accumulated drift errors, we followed a similar RPE evaluation protocol from \cite{qiu2023airimu} as follows:
We divided trajectories into intervals of fixed duration $\Delta t=0.2$\,sec, assigning identical initial conditions (i.e., initial pose and velocity) to all methods at the beginning of each interval, and evaluated the cumulative error over that time period as follows:
\begin{itemize}
\item \text{APE} = $\frac{1}{N} \sum_{i=1}^{N} \left\| \mathbf{p}_{i}^{\text{est}} - \mathbf{p}_{i}^{\text{gt}} \right\|_2$
\item \text{RPE} = $\frac{1}{N} \sum_{i=1}^{N} \left\| \mathbf{p}_{i+\Delta t} - \mathbf{p}_i - \mathbf{R}_i \mathbf{R}_{\text{GT}}^T (\mathbf{p}_{i+\Delta t} - \mathbf{p}_i) \right\|_2$.
\end{itemize}
Here, for APE, $\mathbf{p}_{i}^{\text{est}}$ and $\mathbf{p}_{i}^{\text{gt}}$ represent estimated and ground truth positions. 
For RPE, $\mathbf{p}_i$ and $\mathbf{R}_i$ represent the estimated position and rotation at the interval start $t_i$, $\mathbf{R}_{\text{GT}}$ denotes the ground truth rotation, and $N$ is the number of intervals. 
Both metrics report errors in meters.
{\setlength{\fboxsep}{1pt}
In the result tables, cells were color-coded to indicate performance rankings: \colorbox{myemerald!100}{\textbf{1st}}, \colorbox{mylightgreen!100}{2nd}, \colorbox{myyellow!100}{3rd}. 

\subsubsection{Comparative Methods}
We compared our approach against four methods: a Baseline (i.e., dead-reckoning IMU preintegration) and three state-of-the-art learning inertial odometry methods: TLIO~\cite{liu2020tlio}, AirIMU~\cite{qiu2023airimu}, and AirIO~\cite{qiu2025airio}. 
For fair comparison, we augmented methods without EKF-based frameworks (e.g., AirIMU and Baseline) by adding our LiDAR-based PGO module to address their lack of global drift correction, denoting these augmented variants with an asterisk (*).

To evaluate our motion-aware training strategy's robustness, we deliberately created challenging conditions by limiting training to a single sequence with only 30 epochs, which could potentially lead to overfitting.
These challenging conditions were designed to create potential overfitting scenarios to test our approach.
Under these constrained settings, all learning-based methods were trained equally, except TLIO, which required 100 epochs to achieve meaningful results for comparison. 
All methods employed their publicly available implementations with recommended hyperparameters.
We evaluated three variants of our method (i.e., Ours$^\ddagger$, Ours$^\dagger$, and Ours; see \tabref{tab:inhouse} footer for details).

Note that AirIO\,/\,AirIMU and Ours$^\dagger$\,/\,Ours share identical RPE values within each pair, as they use the same IMU measurement correction networks but differ only in post-processing (i.e., EKF filtering) and inference (i.e., adaptive weighting).
\begin{figure}[t]
    \centering
    \def\width{0.48\textwidth}%
        {%
     \includegraphics[clip, width=0.48\textwidth]{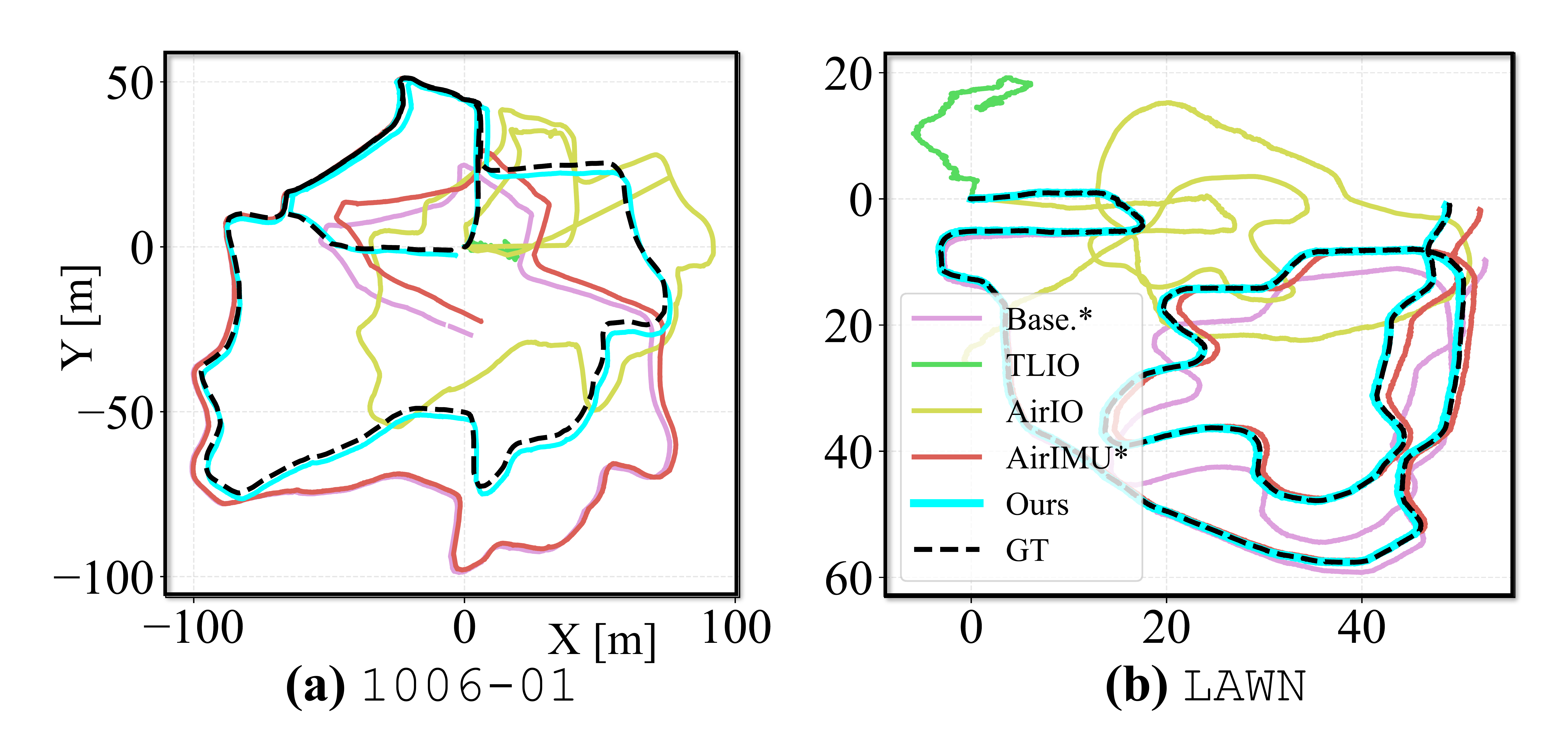}
        }
    \vspace{-0.6cm}
    \caption{Trajectory comparison on unseen sequences (a) \texttt{1006-01} and (b) \texttt{LAWN} showing our method versus comparative approaches.} 
    \label{fig:inhouse_comparison}
    \vspace{-0.2cm}
\end{figure}
\begin{figure}[t]
    \centering
    \def\width{0.4\textwidth}%
        {%
     \includegraphics[clip, width=0.45\textwidth]{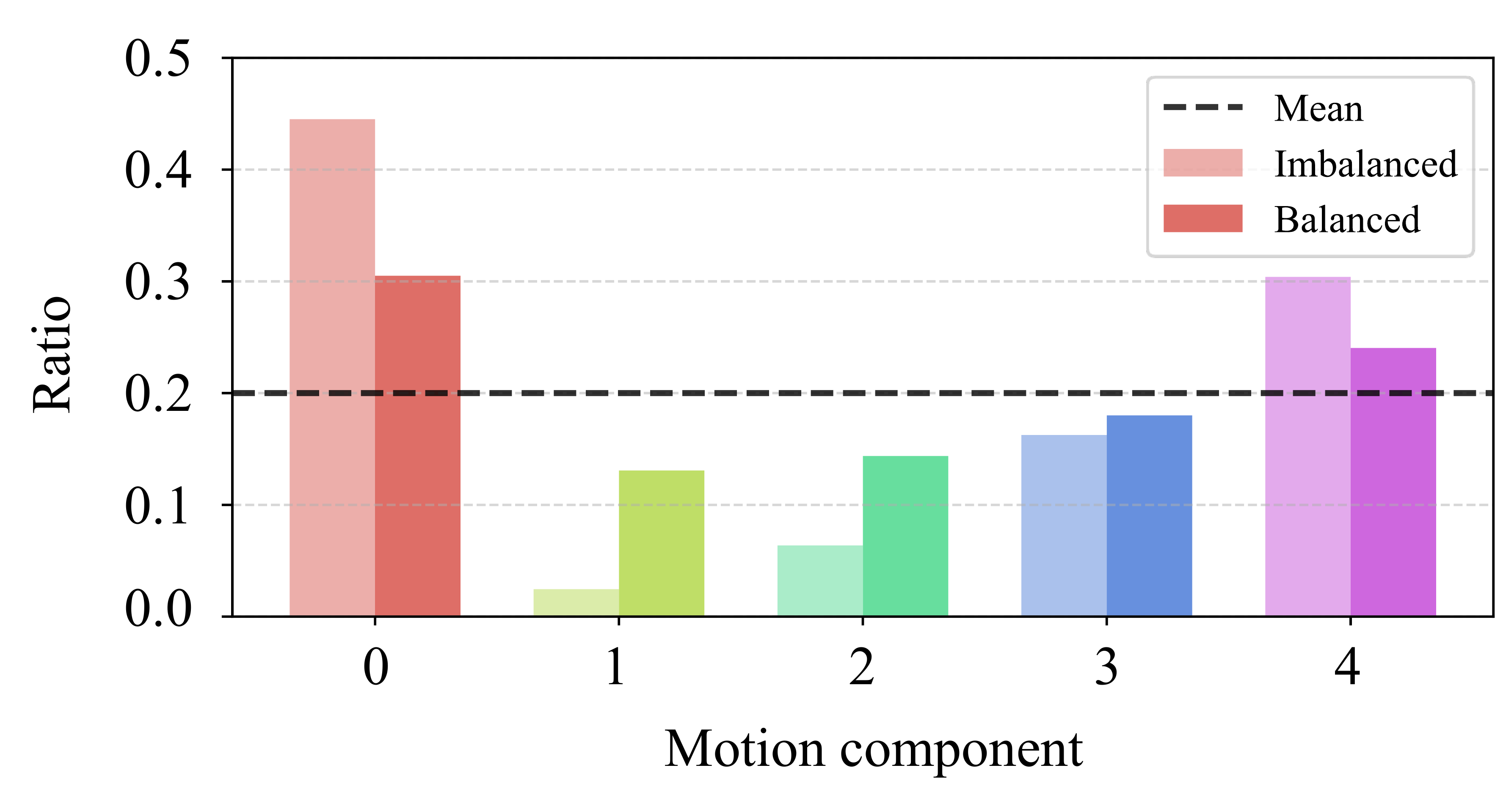}
        }
    \vspace{-0.1cm}
    \caption{Motion component weighting before (imbalanced) and after (balanced) GMM balancing on \texttt{Forest}'s  20\,$\%$ training sequence, demonstrating transition from biased to uniform motion coverage.}
    \label{fig:hist}
    \vspace{-0.6cm}
\end{figure}
\subsection{Performance on Environmental Challenges} 
\tabref{tab:main} shows quantitative comparison results in \texttt{Botanic Garden} dataset and training data percentages. 
Most existing methods, including our method, sometimes maintain lower robustness than the enhanced baseline (Baseline$^*$) on unseen sequences, revealing their tendency to overfit to specific motion patterns and environmental conditions. 
Among them, our approach maintains comparatively robust performance across environments through motion-aware training.
In particular, our approach with adaptive weighting (i.e., Ours) achieves \textit{strong} results with only 20\,$\%$ training data while demonstrating consistent generalization to unseen sequences.
TLIO exemplifies the overfitting problem, showing robust performance on seen sequences with full training data but sharp degradation on unseen sequences as training data decreases.

\subsection{Performance on Motion and Environmental Complexity}

\tabref{tab:main} shows that both Ours and its variants consistently maintain robust performance, ranking between \colorbox{myemerald!100}{\textbf{1st}} and \colorbox{myyellow!100}{3rd} across almost all metrics. 
Notably, the improvement from Ours$^\ddagger$ to Ours$^\dagger$ demonstrates consistent RPE reduction across all sequences, validating the \textit{stable} performance of GMM-based motion analysis.
As shown in \figref{fig:hist}, our approach maintains consistent performance using only 20\% training data, showing that motion diversity is more important than data volume for generalization.
TLIO shows divergent behavior on \texttt{PARK} due to motion distribution differences.
\figref{fig:gmm} shows \texttt{Forest} has a Wasserstein distance of $d_{\text{FtoL}}$\,=\,0.358 with \texttt{LAWN}, while \texttt{PARK} shows $d_{\text{FtoP}}$\,=\,0.604, indicating different motion distributions that lead to performance degradation across most methods.

\begin{table}[t]
\captionsetup{width=0.49\textwidth, justification=justified}
\caption{Feasibility analysis on \texttt{In-House} dataset.
The highly complex motions and featureless environment prevent the acquisition of ground truth, making supervised methods (\rl{\xmark}) infeasible, while our self-supervised approach (\gl{\cmark}) remains deployable.}
\renewcommand{\arraystretch}{1.2}
\centering
\scriptsize
\begin{tabularx}{0.49\textwidth}{l|*{6}{>{\centering\arraybackslash}X}}
\toprule
\hline
Dataset       & \multicolumn{6}{c}{\texttt{In-House} ($\bigstar \bigstar \bigstar \bigstar \bigstar$)} \\ 
\cmidrule(lr){2-7}
Method        & TLIO & AirIO & AirIMU$^{*}$ & Ours$^\dagger$ & Ours$^\ddagger$ & Ours \\ \hline
Feasibility   & \rl{\xmark} & \rl{\xmark} & \rl{\xmark} & \firstc \gl{\cmark} & \firstc \gl{\cmark} & \firstc \gl{\cmark} \\ 
\hline 
\bottomrule
\multicolumn{7}{r}{($\ddagger$) indicates the absence of adaptive weighting and GMM-based motion analysis.} \\
\multicolumn{7}{r}{($\dagger$) indicates the absence of adaptive weighting.} \\
\end{tabularx}
\label{tab:inhouse}
\vspace{-0.4cm}
\end{table}

\subsection{Performance on Extreme Conditions}
To demonstrate the practical advantages of our self-supervised approach, we conducted a feasibility analysis on our \texttt{In-House} dataset, which represents the most challenging conditions. 
As shown in \tabref{tab:inhouse}, the complex terrain renders traditional supervised methods infeasible due to the inability to acquire ground truth data. 
Motion capture systems cannot be deployed in such outdoor environments, and the highly dynamic quadruped motions prevent maintaining the centimeter-level accuracy required for supervised training.
In contrast, our \textit{self-supervised} approach remains fully deployable, requiring only inertial and point cloud data without external ground truth. 
As shown in \figref{fig:steam}, our method maintains consistent mapping even under these extreme conditions.
This highlights the practical value of self-supervised learning for real-world applications where supervised methods face deployment constraints.
\begin{figure}[t]
    \centering
    \def\width{0.48\textwidth}%
        {%
     \includegraphics[clip, width=0.48\textwidth]{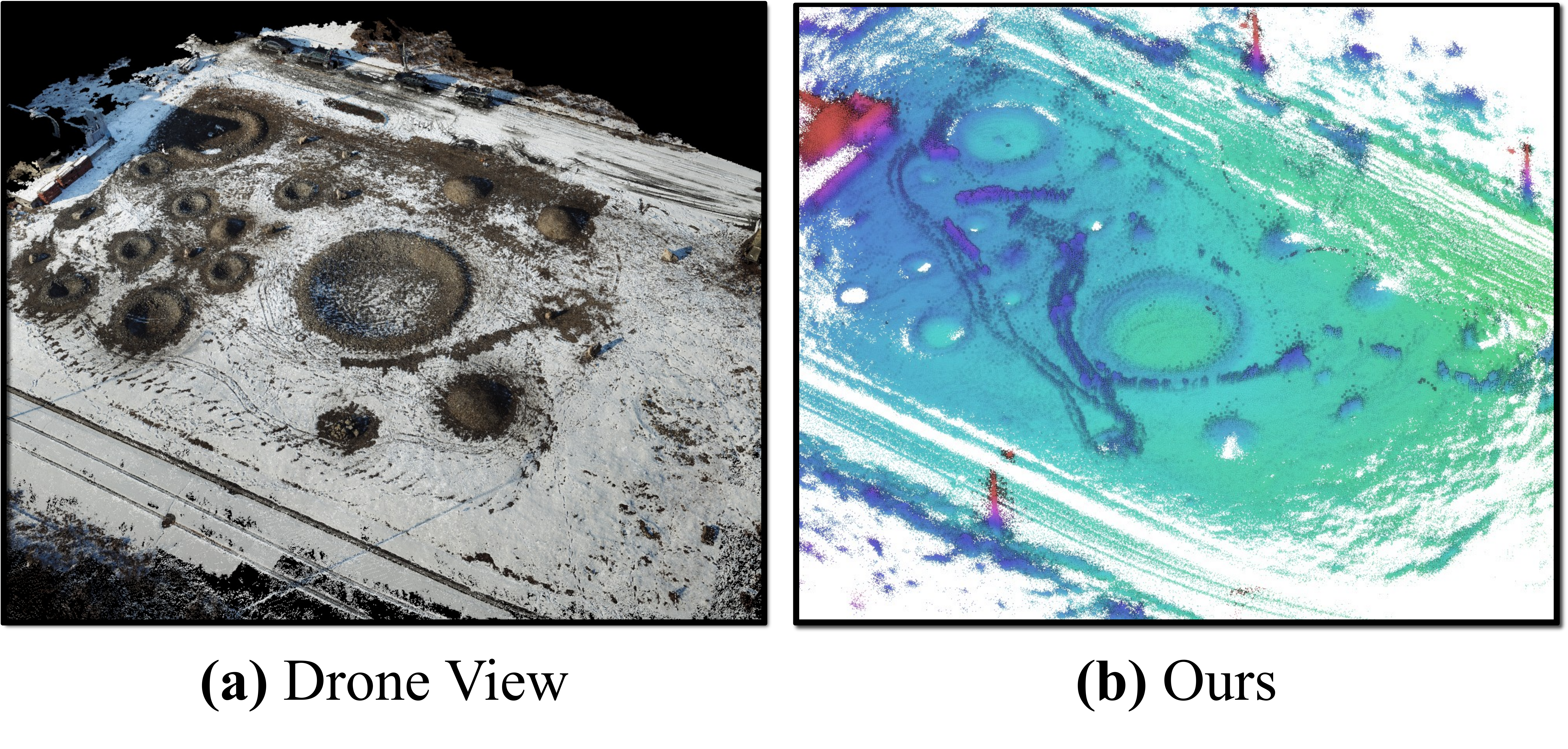}
        }
    \caption{Mapping results comparison on \texttt{In-House} dataset. (a) Aerial view of crater-filled planetary terrain with Unitree B2 quadruped robot. 
         (b) Our mapping results. The proposed method maintains consistent mapping quality despite challenging environmental and motion conditions.}
    \label{fig:steam}
    \vspace{-0.3cm}
\end{figure}
\begin{figure}[t]
    \centering
    \def\width{0.48\textwidth}%
        {%
     \includegraphics[clip, width=0.48\textwidth]{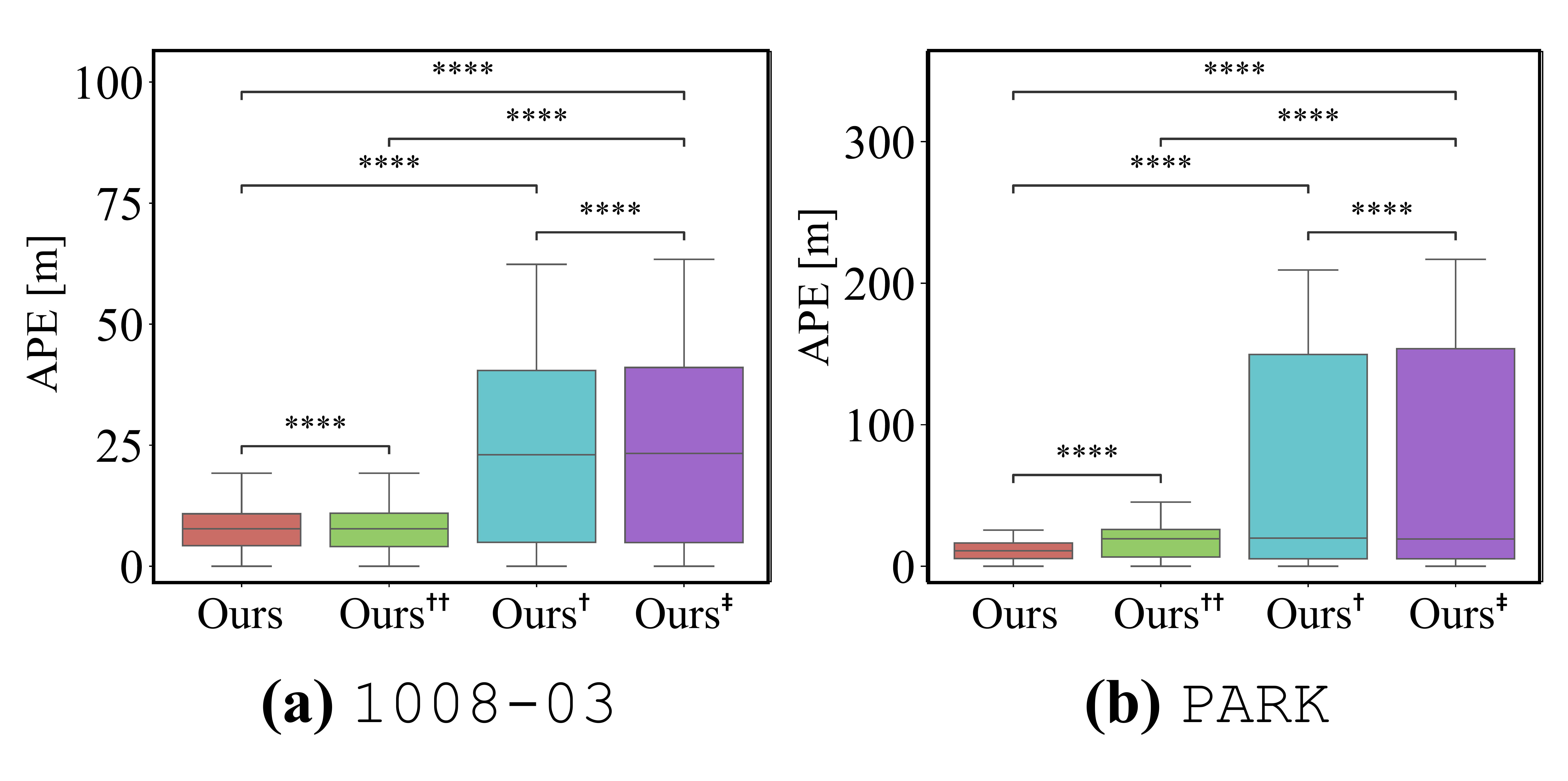}
        }
    \caption{Absolute pose error (APE) box plots on unseen sequences: (a) \texttt{1008-03} and (b) \texttt{PARK}. 
             Statistical significance ($*$$*$$*$$*$: $p$-value $<$ 10$^{-4}$ after a paired $t$-tests) confirms our method's consistent performance improvements.}
    \label{fig:box_plot}
    \vspace{-0.4cm}
\end{figure}

\subsection{Ablation studies}
Finally, we conducted ablation studies across diverse robotic platforms and motion patterns to evaluate the contribution of each component within our framework. 
\figref{fig:box_plot} demonstrates statistically significant performance improvements across all framework components (GMM balancing and adaptive weighting).
This validates the generalizability and effectiveness of our integrated approach through comprehensive statistical analysis.
Notably, Ours$^{\dagger\dagger}$ (without GMM but with adaptive weighting) shows the \colorbox{mylightgreen!100}{2nd} performance benefits of adaptive weighting alone, with detailed trajectory comparisons illustrated in \figref{fig:main}.
\section{Conclusion \& Discussion} \label{Conclusion}
We propose a novel self-supervised inertial odometry (IO) framework, called \textit{KISS-IMU}, that eliminates ground truth dependency through selective LiDAR registration or PGO and GMM-based motion analysis. 
Our approach achieves \textit{stable} IO through balanced motion learning and \textit{strong} IO through adaptive weighting based on uncertainty-aware inference. 
Experimental results demonstrate that our method maintains robust performance even as complexity increases, ranging from environmental challenges in natural settings to highly dynamic quadruped robot motions in challenging terrains
Notably, our framework demonstrates extensibility by successfully integrating with other IO methods, showing its potential for broader applicability beyond the tested implementations.
To the best of our knowledge, this is the first framework that learns only IO in a self-supervised manner, distinguishing it from existing approaches that require either ground truth supervision or learnable supervision. 
Our motion analysis techniques open promising directions for future work, particularly in online learning and test-time adaptation. 


\scriptsize
\bibliographystyle{Packages/IEEEtranN} 
\balance
\bibliography{Packages/string-short, Packages/references}

\end{document}